\documentclass[peerreview]{IEEEtran}
\usepackage{cite} 
\usepackage[utf8]{inputenc} 
\usepackage{booktabs} 
\usepackage[subrefformat=parens,labelformat=parens]{subcaption} 
\usepackage{graphicx}
\usepackage{natbib}
\usepackage{forest}
\usepackage{tabularx}
\usepackage[justification=centering]{caption}
\usepackage{listings}
\usepackage{hyperref}
\usepackage{url} 

\begin{document}
\title{SonoVision: A Computer Vision Approach for Helping Visually Challenged Individuals Locate Objects with the Help of Sound Cues }

\author{
Md Abu Obaida Zishan \\
\and
Annajiat Alim Rasel \\
Department of Computer Science and Engineering\\
BRAC University\\
}

\date{8/9/25}

\maketitle

\begin{center}
  \url{https://github.com/MohammedZ666/SonoVision}
\end{center}
\tableofcontents
\listoffigures
\listoftables

\IEEEpeerreviewmaketitle
\begin{abstract}
Locating objects for the visually impaired is a significant challenge and is something no one can get used to over time. However, this hinders their independence and could push them towards risky and dangerous scenarios. Hence, in the spirit of making the visually challenged more self-sufficient, we present SonoVision, a smart-phone application that helps them find everyday objects using sound cues through earphones/headphones. This simply means, if an object is on the right or left side of a user, the app makes a sinusoidal sound in a user's respective ear through ear/headphones. However, to indicate objects located directly in front, both the left and right earphones are rung simultaneously. These sound cues could easily help a visually impaired individual locate objects with the help of their smartphones and reduce the reliance on people in their surroundings, consequently making them more independent. This application is made with the flutter development platform and uses the Efficientdet-D2 model for object detection in the backend. We believe the app will significantly assist the visually impaired in a safe and user-friendly manner with its capacity to work completely offline. Our application can be accessed here \url{https://github.com/MohammedZ666/SonoVision.git}.
\end{abstract}

\section{Introduction}
\subsection{Background and Motivation}
Visually impaired (VI) individuals have to overcome crucial challenges in their everyday lives. These challenges range from identifying obstacles to their movement, navigation, adapting to difficult weather, and so on. On the other hand these challenges are compounded by the fact that their surroundings are not designed keeping their specific hurdles in mind. Hence, they have to rely on assistive technologies to navigate their environment and relieve some hardship from their life \cite{akilandeswari2022design, kbar2016utilizing, tyagi2021assistive}.  

A crucial part of navigation is object detection. Again, objects are necessary in our everyday life. But for the visually impaired, neither navigation nor finding their necessary objects is a trivial task. Hence, in this work, we leverage the processing power of modern smartphones and their camera to develop a smartphone application that can allow the VI to detect and locate objects precisely with the help of sound cues and CNN based object detection. Our contributions in this work are as follows:

\begin{itemize}
    \item We develop a flutter app to help VI users locating objects they specify using CNN based object detection.
    \item We provide the code-structure, tech-stack, and the working principle of our application.
    \item We discuss the strengths, limitations, and future work for our application.  
\end{itemize}

\subsection{Outline}
In Section \ref{sec:working-methodology} we discuss the working methodology of our application. Section \ref{sec:application-architecure} discusses the application architecture and tech-stack. Section  \ref{sec:cnn-arch} discusses the CNN architecture used in this application and its adaptability of open-set object detectors. Finally, Section \ref{sec:limitations-future-work} discusses future work and concludes the paper. 

\section{Working methodology}
\label{sec:working-methodology}
\begin{figure}[!h]
\centering
\includegraphics[width=1\columnwidth]{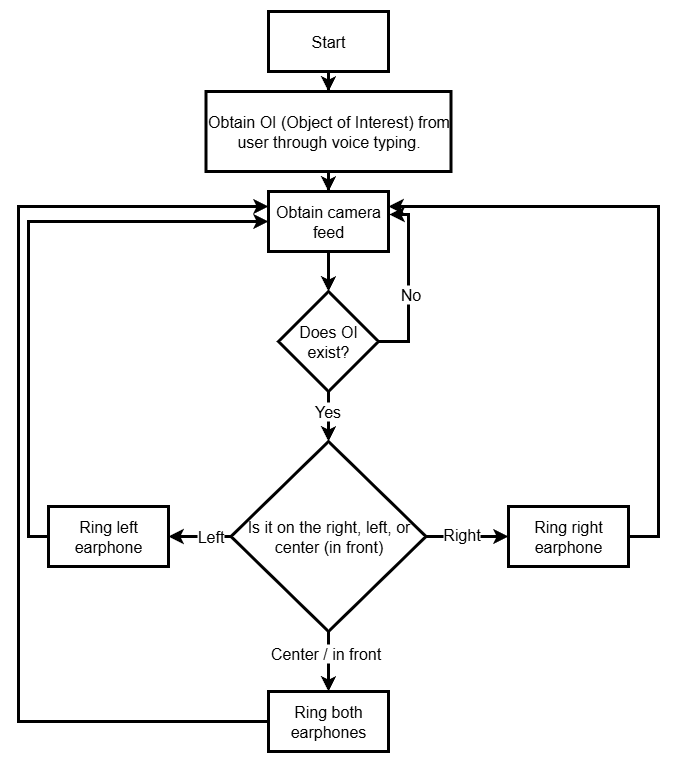} 
\caption{Our application workflow}
\label{fig:flow}
\end{figure}

The application works in the following way, (1) The user inputs his/her object of interest (OI) through voice command to our application. (2) The application runs the object detection model for each frame, delivered from the smartphone's internal Camera API. This returns all objects, the model was trained to detect.(3) Then the application filters detected objects and keeps only one instance of the OI holding the highest confidence. (4) Finally, the application uses sound cues to guide the user toward the OI. (5) If the object is in front of the user, the application rings both ear/headphones of the user for indication. Likewise, if the object is on the left or right of the user, the respective earphone is rung. (6) However, if the object does not exist, the application does not make any sound until an instance of OI falls into frame. This is depicted as a flowchart in Figure \ref{fig:flow}.

\begin{figure*}[ht]
\centering
\begin{subfigure}[t]{0.3\linewidth}    
    \centering
    \includegraphics[width=1\linewidth]{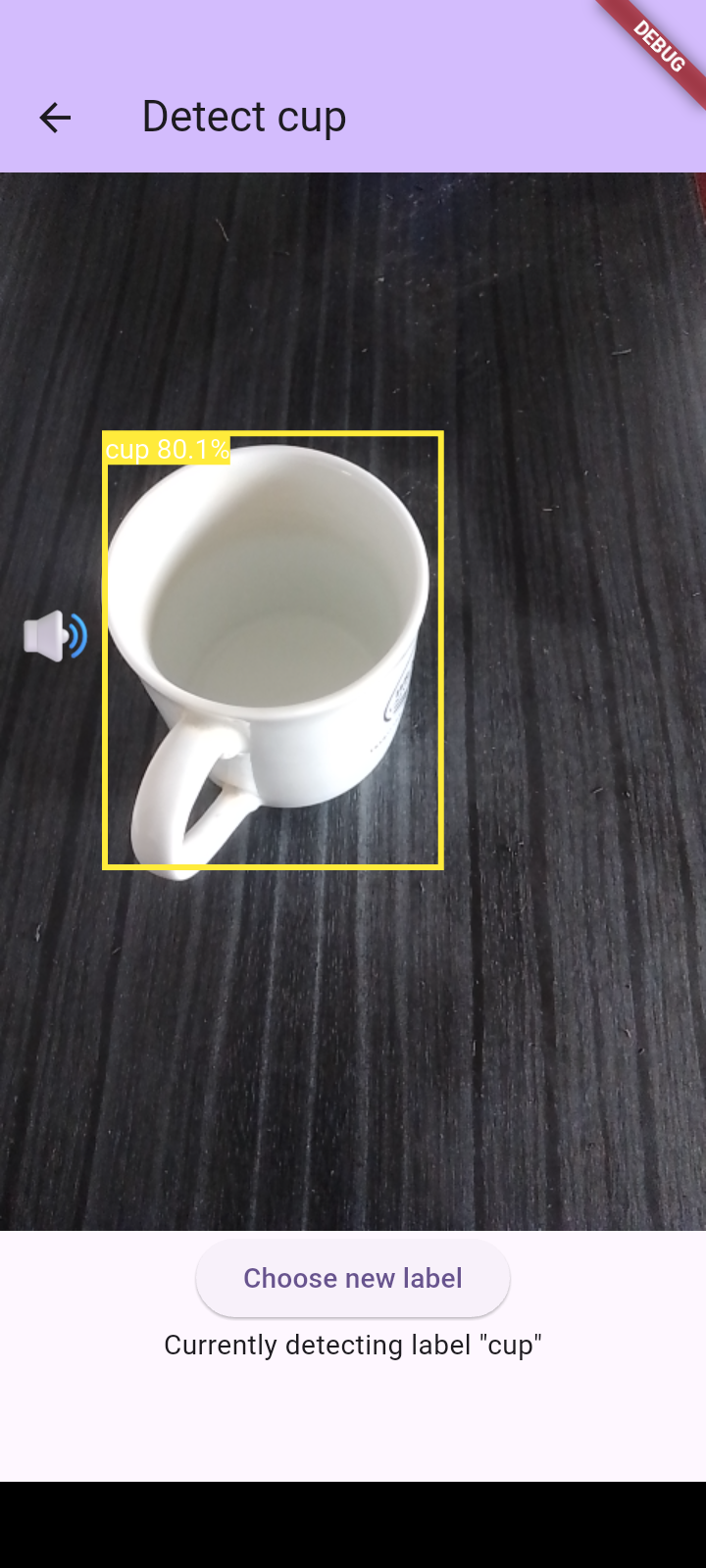}
    \caption{Left ear/headphone ringing to indicate the cup's position on the left.}
    \label{fig:cup_left}        
\end{subfigure}
 \begin{subfigure}[t]{0.3\linewidth}    
 \centering
    \includegraphics[width=1\linewidth]{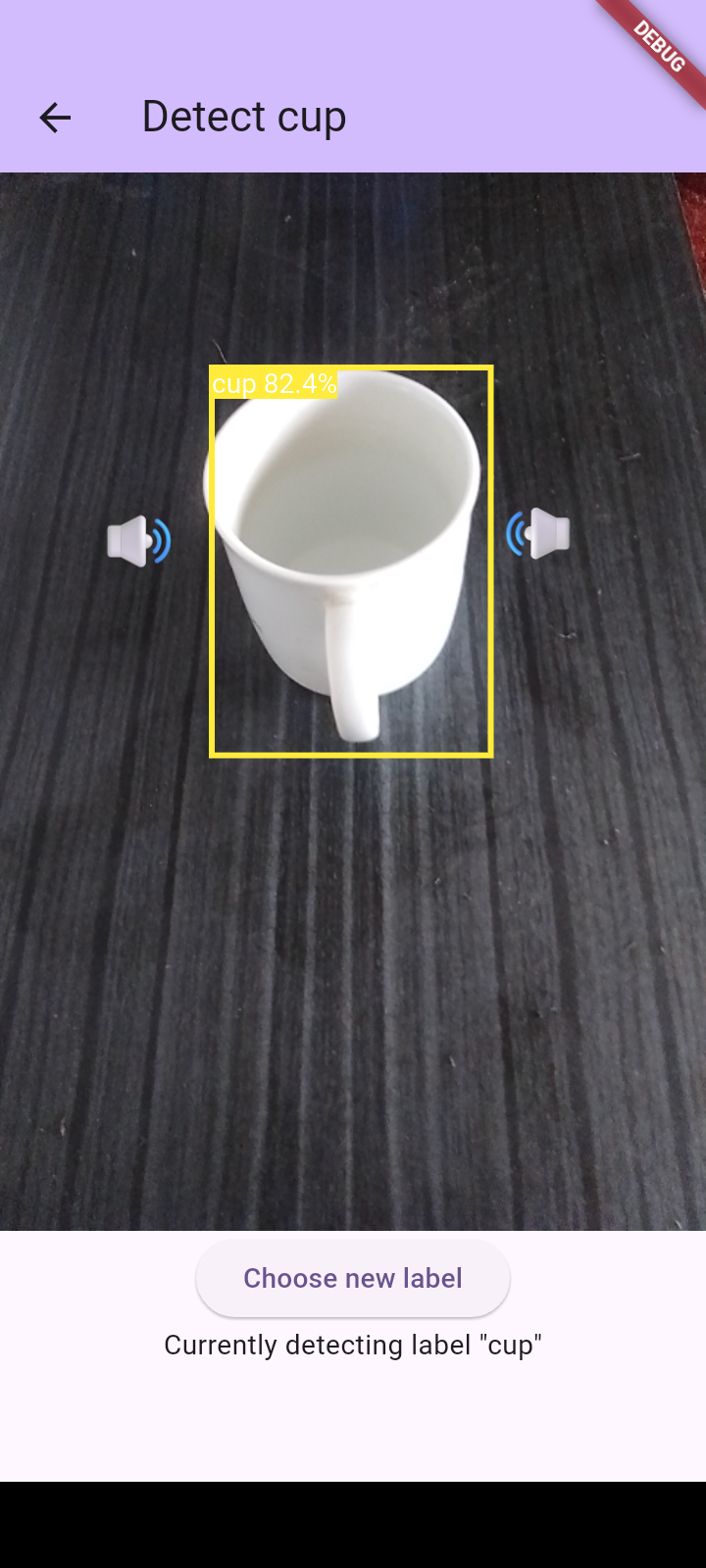}
    \caption{Both ear/headphones ringing the center position of the object.}
    \label{fig:cup_center}
\end{subfigure}
 \begin{subfigure}[t]{0.3\linewidth}    
    \centering
    \includegraphics[width=1\linewidth]{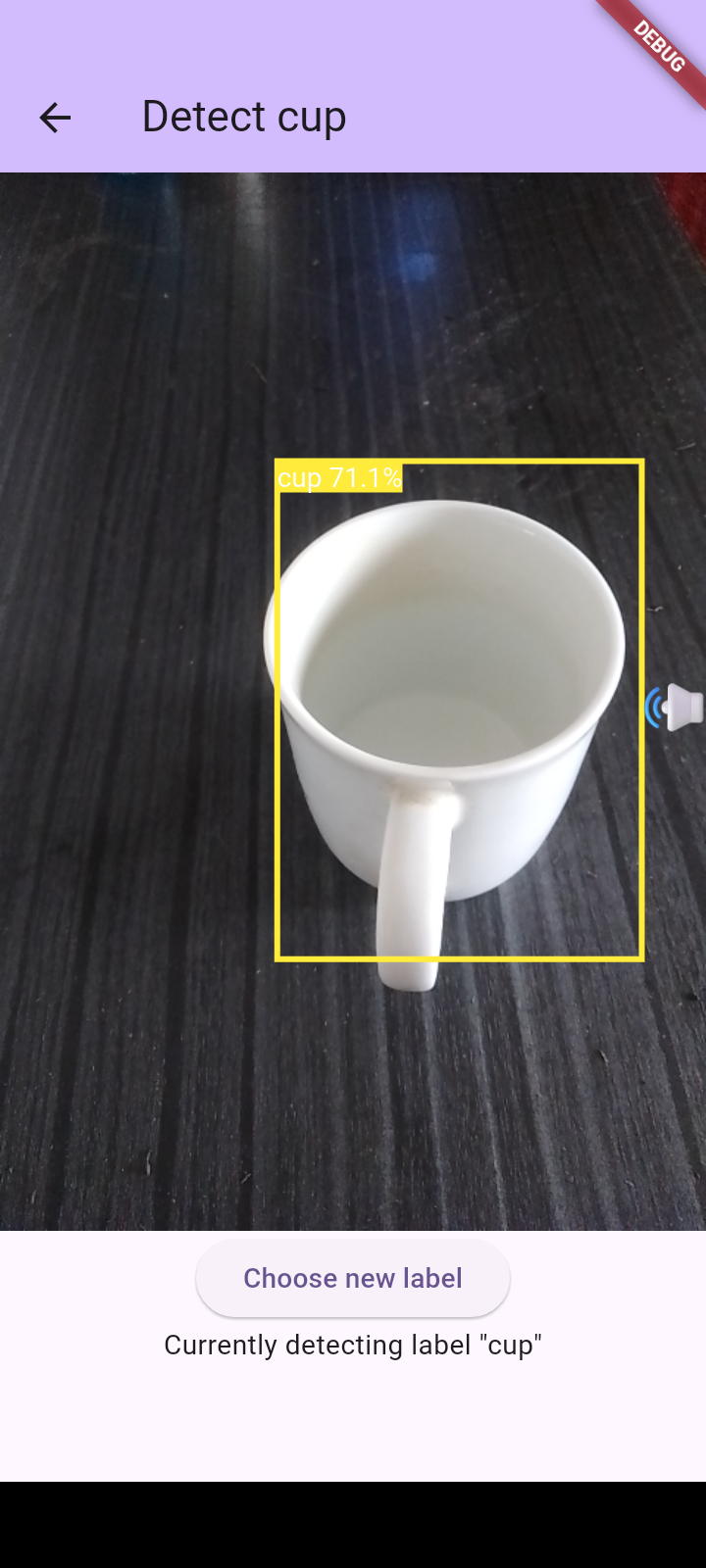}
    \caption{Right ear/headphone ringing to indicate the cup's position on the right.}
    \label{fig:cup_right}        
\end{subfigure}
\caption{Figures \ref{fig:cup_left}, \ref{fig:cup_center}, and \ref{fig:cup_right} shows the working principle of our application, indicating the position of an OI (cup) on left, center, and, right sides, through sound cues in the left, both, and right ears respectively (through ear/headphones). The figures also show the active ear/headphone(s) as indicated by the speaker emoji.} 
\label{fig:application-working-method}
\end{figure*}

Further more, we demonstrate our application in operation through Figure \ref{fig:application-working-method}. This shows the exact method of activating sound cues for the specific ear through head/earphones, once an OI is detected.

\section{Tech Stack \& Application Arhcitecture}
\label{sec:application-architecure}
\subsection{Tech Stack}
To develop our application we chose the flutter development platform. Flutter is a platform-agnostic framework, allowing development of application across all popular platforms, namely-Android, iOS, Desktop, and Browser. We chose this platform due to its versatility and ease-of-development for both front and backend. Hence, developing the application once, we can deploy it to most platforms, due to its platform-specific embedder \cite{flutter_architectural_overview_2025}. 
\subsection{Application Architecture}
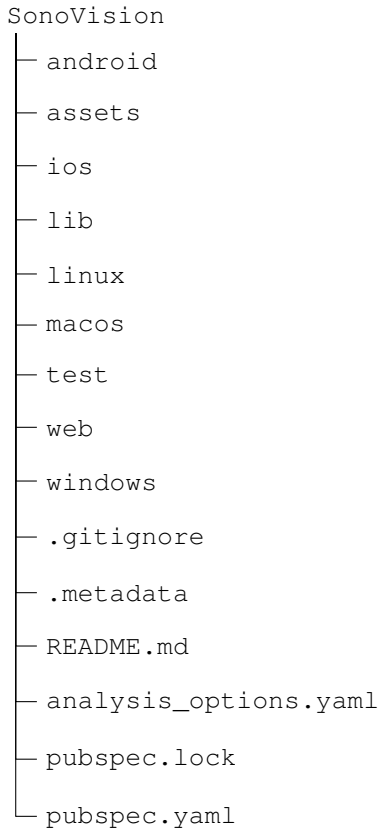
\begin{figure}
\centering
\begin{forest}
for tree={
  font=\ttfamily,
  grow'=0,
  child anchor=west,
  parent anchor=south,
  anchor=west,
  calign=first,
  edge path={
    \noexpand\path [draw, \forestoption{edge}]
    (!u.south west) +(7pt,0) |- (.child anchor) \forestoption{edge label};
  },
  before typesetting nodes={
    if n=1 {insert before={[,phantom]}} {}
  },
  fit=band,
  before computing xy={l=15pt},
}
[SonoVision
  [android]
  [assets]
  [ios]
  [lib]
  [linux]
  [macos]
  [test]
  [web]
  [windows]
  [.gitignore]
  [.metadata]
  [README.md]
  [analysis\_options.yaml]
  [pubspec.lock]
  [pubspec.yaml]
]
\end{forest}
    \caption{Directory structure of our application}
    \label{fig:dir_structure}
\end{figure}

In flutter the starting point of the application is through the main function, usually placed in the \verb|main.dart| file \cite{flutter_intro_dart}. This function and file, along with other helper functions are placed in the \verb|lib| folder. Everything outside the \verb|lib| contains platform specific code as shown in Figure \ref{fig:dir_structure}.    

Due to platform-agnostic development approach of our application, our discussion will be limited to the contents of the \verb|lib|, \verb|assets|, and \verb|pubspec.yaml| folders and file respectively. Rest of the platform specific code, are mostly handled by the flutter development framework. 

\subsubsection{\texttt{lib}}
\begin{figure}
\centering
\begin{forest}
for tree={
  font=\ttfamily,
  grow'=0,
  child anchor=west,
  parent anchor=south,
  anchor=west,
  calign=first,
  edge path={
    \noexpand\path [draw, \forestoption{edge}]
    (!u.south west) +(7pt,0) |- (.child anchor) \forestoption{edge label};
  },
  before typesetting nodes={
    if n=1 {insert before={[,phantom]}} {}
  },
  fit=band,
  before computing xy={l=15pt},
}
[SonoVision/lib
  [models]
  [service]
  [utils]
  [widgets]
  [main.dart]
]
\end{forest}
    \caption{Directory structure of the \texttt{lib} folder}
    \label{fig:dir_structure_lib}
\end{figure}
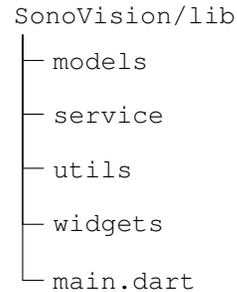

The \verb|lib| folder consists of vital application code that drives the main functionality of this application. The entry point of the application is the \verb|main.dart| file as shown in Figure \ref{fig:dir_structure_lib}. The folder \verb|models| consist of functionalities to handle the output from the CNN model and configuring the application for drawing bounding boxes on screen with the classes \verb|DetectionResult| and \verb|ScreenParams|. Again, \verb|service| consists of the classes \verb|Detector| and \verb|_Detector_Sever|, implementing the functionality of a flutter-isolate. A flutter isolate is the equivalent of a process or a service that communicates with the main application through message passing. This allows the CNN model to process camera-frames in the background for real-time object detection. However, instead of an isolate, if the frames are processed in the foreground, the performance of real-time object detection suffers significantly. The folder \verb|utils| contains the \verb|ImageUtils| class that implements two basic image processing functionalities which are centering a bitmap and converting YUV420 image format to BGRA8888. Finally, the \verb|widgets| folder consists of the implementation of UI components used in this application. These are, (1) \verb|DetectorWidget|, which is responsible for detecting objects and initiating sound cues in the application, (2) \verb|BoundingBoxPainter|, that draws bounding boxes over streamed camera frames, and (3) \verb|SelectionScreen| widget, which allows the user to select his/her OI after the application starts. 

\subsubsection{\texttt{assets}}
The \verb|assets| folder consists of \verb|efficientdet-lite2.tflite| file which is the tflite version of our CNN model. The model weights are loaded from this directory during the application's operation.

\subsection{\texttt{pubsec.yaml}}
This file contains the flutter packages used in this application for building all functionalities. The flutter CLI tool automatically builds the application for debug or release using the packages mentioned in this file. The most crucial packages we used were: 
\begin{enumerate}
    \item (1) \verb|tflite_flutter| \cite{tflite_flutter}: This package allows the inference of our CNN model through the Tensorflow API. It also includes CPU (XNNPACK \cite{xnnpack}) and GPU optimizations for the inference of neural networks
    \item \verb|flutter_soloud| \cite{flutter_soloud}: This package facilitates the functionality of sound cues through earphones
    \item \verb|camera|: This package is used for streaming the camera-feed from the device's built-in camera to the application
\end{enumerate}

Besides these packages, other flutter packages were also used for improving the user-interface and experience. This concludes our application's architectural overview and discussion. In the next section, we will discuss the architecture of our pretrained CNN model and flexibility of choosing a different model.

\section{Architecture, Limitations, and Adaptability of Pretrained Object Detection Models}
\label{sec:cnn-arch}
We developed this application in a modular manner. Hence, swapping the pretrained CNN architecture, simply by choosing a different model is trivial. Therefore, in this section we will discuss, the current CNN architecture in use and its scope, and the methods for selecting a different architecture with more robust object detection and classification performance for future work. 

\subsection{EfficientDet-D2}
\begin{figure*}
    \centering
    \includegraphics[width=1\linewidth]{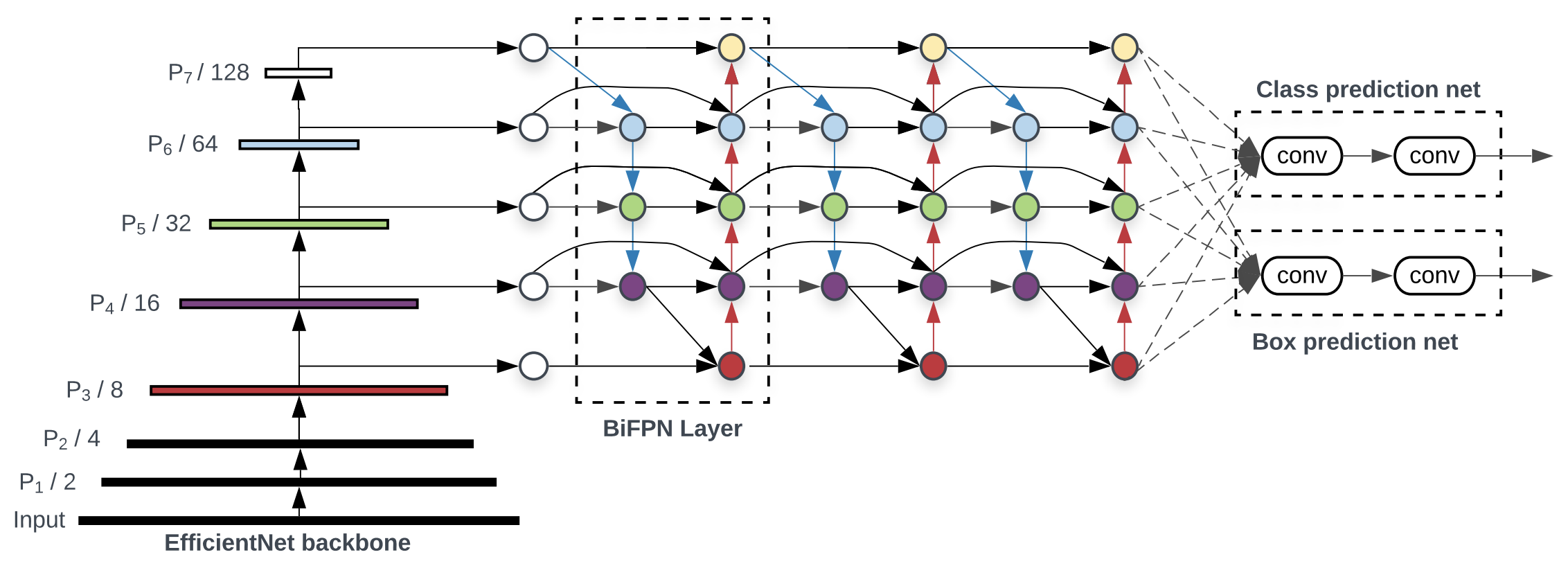}
    \caption{Network architecture of EfficientDet model family}
    \label{fig:efficientdetstruct}
\end{figure*}

EffcientDet is a family of weighted bi-directional feature pyramid networks achieving 34.6-55.1 mAP on COCO (Common Objects in Context) dataset (test-set) for object detection and classification \cite{tan2020efficientdet}. It consists of 3.9-77M parameters and FLOPs (floating point operations per second) ranging from 2.45B-410B. For our application, we chose the EfficientDet-D2. We state its performance metrics and the general structure of EfficientDet family in Table \ref{tab:metrics} and Figure \ref{fig:efficientdetstruct} respectively.

\subsection{Drawbacks:}
For our application, we require a VI individual to be able to locate any objects in the environment around him/her. However, since the EfficientDet model family was trained on the COCO dataset with only 80 classes, the application falls short on detecting \textit{anything} the user may want to look for. As most popular model families such as YOLO and EfficientDet are trained for closed set object detection. For our application, we require an open-set object detection model, that will be able to detect most objects present in the environment, if not all of them \cite{Zheng_2022_CVPR}. To overcome this drawback, we suggest the using models like GroundingDINO \cite{liu2024grounding}. OWL-ViT \cite{owl-vit}, GLIP \cite{glip} that are trained with open-set object detection in mind and is crucial for our use-case.  

\begin{table*}[h]
\large
\centering
    \begin{tabular}{|c|c|}
        \hline
        Metric & Value \\
        \hline
         $AP_{test}$ & 43.9 \\
         $AP_{50}$ & 62.7 \\
         $AP_75$ & 47.6 \\
         $AP_S$ & 22.9 \\
         $AP_M$	& 48.1 \\	
         $AP_L$ & 59.5 \\				
         $AP_val$ & 43.5 \\		
         $Paramerters$ & 8.1M \\		
         $FLOPs$ & 11.0B	\\
         \hline
    \end{tabular}
    \caption{Performance of EfficientDet-D2 model}
    \label{tab:metrics}
\end{table*}

\subsection{Adaptability}
Since our application was developed keeping open-set object detectors in mind, we can readily adapt to them, once they become efficient enough for smartphones. We only need to change the model file, and rewrite some functionality regarding scaling bounding boxes to adapt to model output (if the model output format differs from EfficientDet family). In this way, our application can readily adopt detecting and locating potentially \textit{everything} for safe, sound, and comfortable livelihoods of VI individuals. 

\section{Conclusion and future work}
\label{sec:limitations-future-work}
with the fast advent of modern computer vision, robotics, and, pattern recognition systems, it was expected that the life of physically and mentally challenged individuals will get exponentially easier. However, we far from keeping the promise. Hence, it is vital that we try our hardest to develop tools that comes in the aid of all challenged individuals. 

Hence, in this paper, we described our developed application that helps visually challenged individuals navigate around their surrounding environment, and locate objects with the help of our application. We discussed the limitations of our system and its capacity to seamlessly integrate a broader open-set object detector for overcoming detecting only a few object-classes. 

Thus, future researchers can focus on developing a real-time open-set object detector with parameters less than 10M for operating on smartphones. Again, modifying existing open-set detectors for smartphones and edge devices using knowledge distillation \cite{gou2021knowledge} and quantization \cite{hubara2018quantized} will also help this field of work significantly. 

\section*{Acknowledgment}
This work was supported in part by the BRAC University Research Seed Grant (RSGI) for \textit{AI-powered Enhancement of University Life Accessibility for Blind People.}

\appendices

\bibliographystyle{unsrt} 
\bibliography{references} 


\end{document}